\documentclass[letterpaper]{article} 
\usepackage{aaai2027}  
\usepackage[hyphens]{url}  
\usepackage{graphicx} 
\usepackage{natbib}  
\usepackage{caption} 
\usepackage{booktabs}

\usepackage{amsmath}
\usepackage{enumitem}
\usepackage{multirow}
\usepackage{xcolor}
\usepackage{colortbl}
\definecolor{tabgood}{RGB}{120,100,220}
\definecolor{tabbad}{RGB}{190,90,70}
\definecolor{tabhead}{RGB}{236,236,236}
\usepackage[most]{tcolorbox}
\definecolor{phc}{HTML}{BE5A46}
\newtcolorbox{promptbox}[2][]{enhanced, colback=black!2, colframe=black!78, coltitle=white,
  fonttitle=\footnotesize\bfseries, title={#2}, boxrule=0.6pt, arc=2pt,
  left=5pt, right=5pt, top=4pt, bottom=4pt, toptitle=2pt, bottomtitle=2pt, middle=4pt,
  segmentation style={black!40, dashed}, fontupper=\footnotesize\ttfamily, fontlower=\footnotesize\ttfamily,
  halign=left, halign lower=left, #1}
\newcommand{\promptpart}[1]{{\normalfont\footnotesize\bfseries\color{black!60}#1}\par\smallskip}
\newcommand{\ph}[1]{\textcolor{phc}{\{#1\}}}
\usepackage{pifont}
\definecolor{markgreen}{HTML}{2F8F4E}
\definecolor{markred}{HTML}{BE5A46}
\newcommand{\cmark}{\textcolor{markgreen}{\ding{51}}}
\newcommand{\xmark}{\textcolor{markred}{\ding{55}}}

\newcommand{\jeticon}{\raisebox{-0.2em}{\includegraphics[height=1.15em]{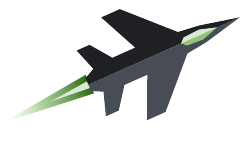}}}
\title{\jeticon\,JET: \enspace Just-Enough Exploration Tuning for Efficient Tool-Calling Agents \\ in Repository-Level Question Answering}

\author{
    Yoonho Choi,
    Taehyeon Han,
    Seonghyeon Lee\corresponding
}
\affiliations{
    School of Computer Science and Engineering, Kyungpook National University, Daegu, South Korea\\
    \{yoonhoc, gksxogus1946, sh0416\}@knu.ac.kr
}
\nocopyright

\begin{document}

\maketitle
\begingroup
\renewcommand{\thefootnote}{}
\footnotetext{Preprint.}
\endgroup
\begin{abstract}
In repository-level question answering (QA), an~agent explores a~code repository with tools and should answer a~question once enough information is gathered.
Nevertheless, we find that LLM agents often continue to invoke tools even after the~gathered information is already sufficient to answer.
We call this phenomenon \emph{over-exploration}.
Moreover, even though existing methods improve task performance through supervised fine-tuning (SFT) on expert trajectories, SFT also propagates the~over-exploration in the~trajectories to the~agent.
To mitigate over-exploration, we propose \textbf{J}ust-\textbf{E}nough Exploration \textbf{T}uning (\textbf{JET}), a~fine-tuning method that trains the~agent to answer once the~gathered information is sufficient.
JET truncates each training trajectory at the~step where the~gathered information becomes sufficient, inserts a~rationale that explains why the~agent can answer at that step, and fine-tunes the~agent with a~higher loss weight on the~rationale.
Experiments on two repository-level QA benchmarks with two open-weight 8B models demonstrate that JET reduces tool calls by 11 to 33\% on self-generated trajectories while largely preserving the~score.
On expert trajectories, JET makes 18 to 28\% fewer tool calls than SFT while retaining most of the~score gain of SFT.
In both cases, JET also lowers the~inference cost.
\end{abstract}


\section{Introduction}
\label{sec:intro}

In software engineering, large language models (LLMs) have evolved from function-level code generation~\citep{chen2021evaluatinglargelanguagemodels} to agents~\citep{yao2023react} that explore repositories with tools and resolve real-world GitHub issues~\citep{jimenez2024swebench, yang2024sweagent}.
Recent work has also studied repository-level question answering (QA), where the~agent answers natural language questions about a~repository~\citep{peng-etal-2026-swe, cai-etal-2026-swe}.

\begin{figure}[!t]
\centering
\includegraphics[width=\columnwidth]{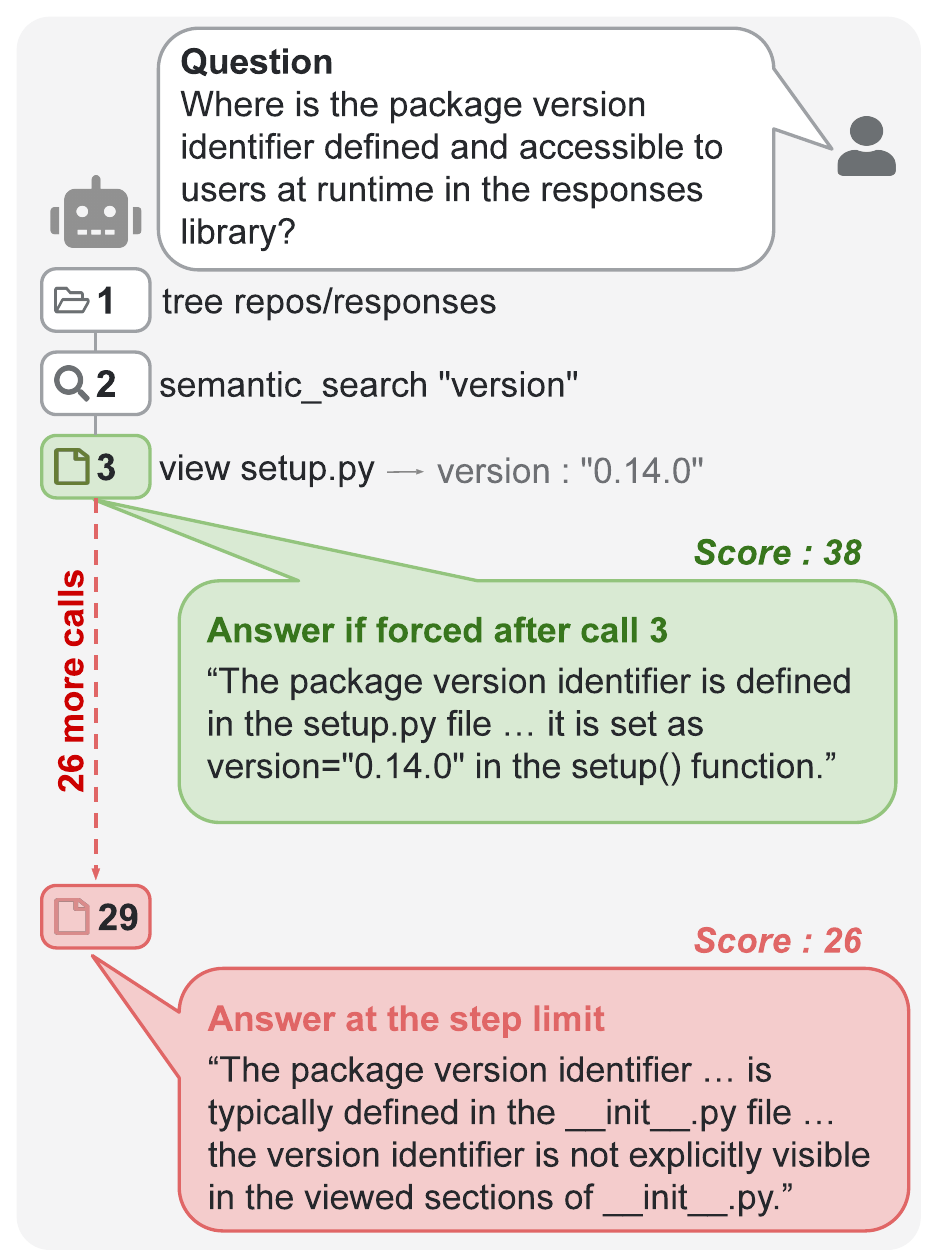}
\caption{An~example of over-exploration on SWE-QA-Pro (Qwen3-8B). The~answer forced after the~third tool call (green) scores higher than the~final answer after 26 more calls (red).}
\label{fig:case}
\end{figure}

\begin{figure}[t]
\centering
\includegraphics[width=\columnwidth]{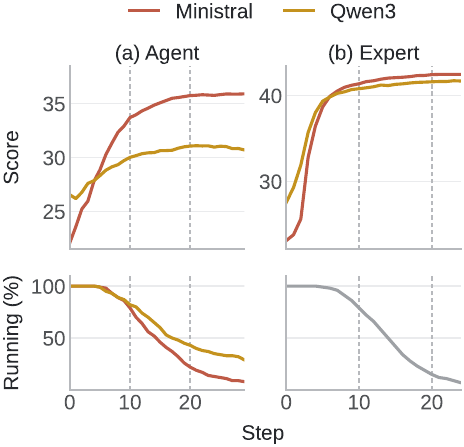}
\caption{Forced-answer score (top) and percentage of trajectories still exploring (bottom) at each step, for (a) agents on the SWE-QA-Pro benchmark and (b) expert trajectories in the SFT training data. In (b), the agents answer from the context of the expert. The score reaches a plateau while the exploration continues. GPT-6 Luna scores the answers.}
\label{fig:waste}
\end{figure}

In repository-level QA, an agent interacts with the repository over a sequence of steps.
At each step, the agent either invokes a tool to explore the repository or writes the final answer.
The agent must therefore decide when to stop exploring and answer the question.
However, we observe that agents often invoke tools excessively, even after gathering sufficient information to answer the question.
As illustrated in Figure~\ref{fig:case}, the agent discovers the relevant file by the third step.
Instead of answering, the agent continues to explore until the step limit is reached.

We call this phenomenon~\emph{over-exploration}, in which the~agent continues to make tool calls after the~gathered information is already sufficient to answer.
To examine over-exploration, we force the~agent to answer after every step and score each intermediate answer.
The~results on SWE-QA-Pro, a~repository-level QA benchmark, show that the~score of the~intermediate answers reaches a~plateau while the~agents often continue to explore (Figure~\ref{fig:waste}a).
Over-exploration increases computational cost, because every tool call appends the~tool output to the~context, and the~agent reprocesses the~entire context at every step.
Moreover, supervised fine-tuning (SFT) on expert trajectories is a~common practice for building open-source agents~\citep{zeng-etal-2024-agenttuning, pan2025training}.
Expert trajectories are high-quality trajectories generated by a~stronger LLM, collected to improve task performance regardless of whether each tool call is necessary.
As a~result, the~score of the~intermediate answers generated from the~expert trajectories also reaches a~plateau while the~exploration continues (Figure~\ref{fig:waste}b).
Since SFT trains the~agent to imitate every step of the~trajectory, the~over-exploration in the~trajectory is distilled into the~agent.

Prior work on efficient tool use falls into inference-time and training methods.
At inference time, Budget Tracker~\citep{liu2026budgetaware} states the~tool-call budget in the~prompt, and an~LLM autorater~\citep{ICLR2025_33dffa2e} decides whether the~context is sufficient to answer.
Our experiments demonstrate that these methods are either ineffective or costly in repository-level QA.
Among training methods, AutoSearch~\citep{sun-etal-2026-autosearch} and DAS~\citep{10.1145/3774904.3792235} force the~agent to answer at intermediate steps and train the~agent to answer as soon as the~intermediate answer is correct.
However, both methods focus on short-answer tasks with fewer than 5 tool calls per question on average, where a~gold answer provides an~explicit binary signal of correctness.
Repository-level QA is more realistic: exploration is long-horizon with more than 10 tool calls per question on average, and a~free-form answer is scored on a~scale.
The~judge score is a~qualitative assessment rather than a~match against a~gold answer, and provides only an~implicit signal of correctness.
Existing training methods are therefore difficult to apply to realistic environments.
To our knowledge, no prior work trains an~agent to decide when to answer directly from the~judge scores of free-form answers in long-horizon tasks.

To mitigate over-exploration in repository-level QA, we introduce Just-Enough Exploration Tuning (JET), which trains the~agent to answer once the~gathered information is sufficient.
JET truncates each training trajectory at the~step where the~gathered information becomes sufficient and inserts a~rationale, which explains why the~information gathered so far is sufficient to answer.
The~agent is then fine-tuned with a~higher loss weight on the~rationale to emphasize the~decision to answer during training.

In our experiments, JET mitigates over-exploration, substantially reducing tool calls while largely preserving the~score on self-generated trajectories.
On expert trajectories, JET retains most of the~score gain of SFT, and even exceeds SFT with Ministral, with 18 to 28\% fewer tool calls.
Every baseline, training-free or fine-tuned, fails to reduce tool calls or inference cost in at least one setting.

Our contributions are summarized as follows.
\begin{itemize}[leftmargin=*,itemsep=2pt]
\item In repository-level QA, we find over-exploration in both agents and the~expert trajectories for SFT, where the~score reaches a~plateau while the~exploration continues.
\item We propose JET, a~fine-tuning method that mitigates over-exploration in tasks where free-form answers are evaluated on a~scale, which provides only an~implicit signal of correctness.
\item We demonstrate that JET not only reduces tool calls and inference cost on both types of training trajectories across all combinations of models and benchmarks, but also largely preserves task performance.
\end{itemize}

\section{Related Work}
\label{sec:related}

\paragraph{LLM agents for code repositories.}
LLM agents now work on entire code repositories for tasks such as resolving GitHub issues~\citep{jimenez2024swebench, yang2024sweagent}.
Repository-level QA benchmarks consist of natural language questions that require code from multiple files of a~repository to answer~\citep{peng-etal-2026-swe, cai-etal-2026-swe}.
To improve task performance, SFT on expert trajectories is widely used in repository-level tasks~\citep{pan2025training, cai-etal-2026-swe} and other agent tasks~\citep{zeng-etal-2024-agenttuning}.
The objective of these works is task success, not the efficiency of tool use in the expert trajectories.

\paragraph{Efficient tool use.}
At inference time, existing methods control tool use via the prompt or an additional LLM.
Budget Tracker appends the used and remaining tool-call budget to the prompt, so that the agent adapts tool use to the remaining budget~\citep{liu2026budgetaware}, and the sufficient-context autorater uses a frontier LLM to judge whether a retrieved context is sufficient~\citep{ICLR2025_33dffa2e}.
However, the remaining budget does not reflect the gathered information, and the autorater requires an extra LLM call at inference time.

Training methods follow two lines of research.
First, some methods decide whether a tool is needed from the parametric knowledge of the model~\citep{qian-etal-2025-smart, xu-etal-2025-alignment, ICLR2026_85255cf8}, not from the information accumulated over the trajectory.
Second, other methods use reinforcement or preference learning to reduce tool calls, such as OTC~\citep{wang2025actingreasoningmoreteaching}, ET-Agent~\citep{chen-etal-2026-et}, DORA~\citep{qin-etal-2026-dora}, and Tool-Light~\citep{ICLR2026_788f0d33}, but provide no step-level supervision on when the information becomes sufficient.
Most closely related to our work are AutoSearch and DAS, which target search agents in agentic RAG~\citep{sun-etal-2026-autosearch, 10.1145/3774904.3792235}.
Both methods generate an intermediate answer at each search step and consider the retrieved information sufficient once the intermediate answer matches the gold answer, which yields a binary signal.

However, the training methods above, including AutoSearch and DAS, target tasks that require short answers, where a gold answer provides explicit supervision.
AutoSearch and DAS make fewer than 5 tool calls per question on average.
In repository-level QA, the agents make more than 10 tool calls per question on average, and free-form answers are scored on a scale, so the supervision is only implicit.
JET differs by determining the step at which the gathered information becomes sufficient from the implicit signal of the scores of free-form answers (Table~\ref{tab:related}).

\begin{table}[t]
\centering
\small
\setlength{\tabcolsep}{4pt}
\newcommand{\rowline}{\arrayrulecolor{black!35}\specialrule{0.3pt}{1.5pt}{1.5pt}\arrayrulecolor{black}}
\begin{tabular}{@{}lccc@{}}
\toprule
\textbf{Method} & \textbf{Free-form} & \textbf{Trained} & \textbf{Sufficiency} \\
\midrule
Budget Tracker & \cmark & \xmark & \xmark \\
\rowline
Autorater & \cmark & \xmark & \cmark \\
\rowline
OTC, ET-Agent, & \multirow{2}{*}{\xmark} & \multirow{2}{*}{\cmark} & \multirow{2}{*}{\xmark} \\
DORA, Tool-Light & & & \\
\rowline
AutoSearch, DAS & \xmark & \cmark & \cmark \\
\rowline
\textbf{JET (ours)} & \cmark & \cmark & \cmark \\
\bottomrule
\end{tabular}
\caption{Comparison with existing methods. \emph{Free-form}: supports free-form answers rated on a scale. \emph{Trained}: trains the agent, without extra inference cost. \emph{Sufficiency}: targets the step at which the gathered information is sufficient.}
\label{tab:related}
\end{table}

\section{Problem Formulation}
\label{sec:problem}

\subsection{LLM Agents for Repository-Level QA}
\label{sec:agents}

In repository-level QA, given a natural language question $q$ about a repository $\mathcal{R}$, the~task is to generate an answer $a$.
The question asks how, where, or why a feature is implemented in the repository, and the answer is a free-form explanation that cites code across multiple source files.
However, $\mathcal{R}$ is much larger than the context window of LLMs.
For this reason, an LLM agent must use tools to explore $\mathcal{R}$ and read only the relevant files.

The agent consists of a language model $\pi_\theta$, which generates every output $y$, and a ReAct-style scaffold~\citep{yao2023react}, which executes the tool calls and controls the loop.
At each step, $\pi_\theta$ generates reasoning, a tool call, or both.
If the step contains a tool call, the scaffold executes the tool call and appends the observation to the context before the next step.
When $\pi_\theta$ decides to answer, $\pi_\theta$ writes the final answer inside a \texttt{<finish>} block, and the scaffold terminates the agent.
If the agent reaches the step limit, the scaffold forces $\pi_\theta$ to answer and terminates the agent.
Step $i$ consists of the output $y_i$ of $\pi_\theta$ and the observation $o_i$.
At each step, $\pi_\theta$ takes the context as input.
After $i$ steps, the context $s_i$ contains the question $q$ and all steps,
\begin{equation}
    s_0 = (q), \quad s_i = (q, (y_1, o_1), \ldots, (y_i, o_i)). \label{eq:context}
\end{equation}
Given $s_i$, $\pi_\theta$ generates the next output $y_{i+1} \sim \pi_\theta(\cdot \mid s_i)$.
Let $n$ denote the number of steps before the agent answers.
The output $y_{n+1}$ then contains the final answer $a$.
An LLM judge evaluates $a$ against a reference answer $a^{\text{ref}}$ and assigns a score $S(a, a^{\text{ref}})$.
The goal is to generate an answer that maximizes $S(a, a^{\text{ref}})$.
At the same time, the inference cost of each step increases with $i$, since $s_i$ contains all previous steps (Eq.~\ref{eq:context}).
An efficient agent should therefore answer once $s_i$ is sufficient to answer the question.

\subsection{Over-Exploration in Repository-Level QA}
\label{sec:overexploration}

In practice, agents often explore beyond the step at which $s_i$ becomes sufficient.
To identify this step, we force the agent to answer from every $s_i$ and score each intermediate answer against the reference answer $a^{\text{ref}}$.
We define the \emph{golden step} $i^\star$ as the earliest step with the highest score, and consider the gathered information sufficient at $i^\star$.
The tool calls after $i^\star$ are unnecessary, because the information needed for the highest score has already been gathered.
We formally refer to these tool calls as \emph{over-exploration}.

On the SWE-QA-Pro benchmark, the average score of the intermediate answers reaches a plateau while the agents continue to explore on many questions (Figure~\ref{fig:waste}a).
The score also reaches a plateau on the expert trajectories for SFT, where each agent answers from the context of the expert and the final answer of each trajectory serves as $a^{\text{ref}}$ (Figure~\ref{fig:waste}b).
The tool calls after the plateau are costly, since the total number of processed tokens grows superlinearly with~$n$.
Motivated by these observations, we investigate \textbf{how to build an efficient agent by mitigating over-exploration}.
\begin{figure*}[t]
\centering
\includegraphics[width=\textwidth]{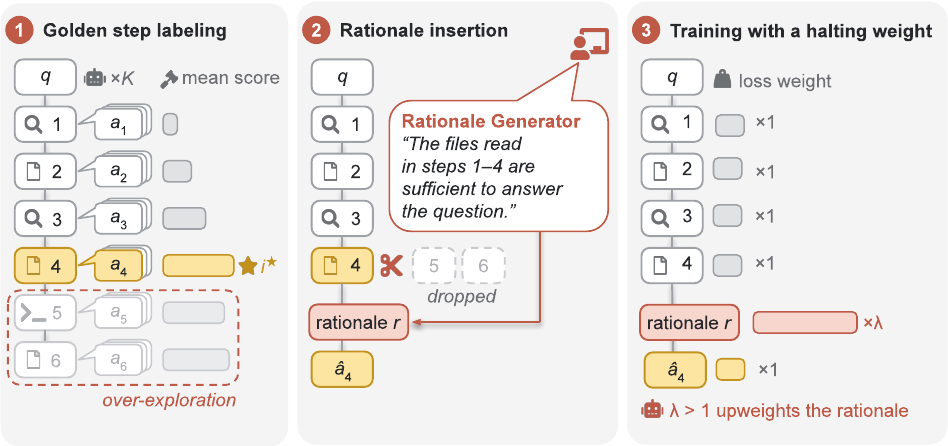}
\caption{Overview of JET. (1) The agent answers $K$ times after every step, and the earliest step with the highest mean score is the golden step $i^\star$. (2) We truncate the trajectory at $i^\star$ and append a rationale from a rationale generator, followed by the golden step answer. (3) We fine-tune the agent with the halting weight $\lambda$ on the rationale.}
\label{fig:overview}
\end{figure*}
\section{Method}
\label{sec:method}

To build an efficient agent with less over-exploration, we introduce \textbf{J}ust-\textbf{E}nough Exploration \textbf{T}uning \textbf{(JET)}, a fine-tuning method that trains the agent to answer once the gathered information is sufficient, rather than to imitate every tool call of the trajectory.
JET consists of three stages, which are golden step labeling, rationale insertion, and training with a halting weight.
Figure~\ref{fig:overview} provides an overview of JET.
We apply JET to both self-generated trajectories and expert trajectories, which we refer to as the Self and Expert settings, respectively.
At inference time, the trained agent itself decides when to answer, without any external signal.

\subsection{Golden Step Labeling}
\label{sec:golden-step}

We force the agent to answer at every $s_i$ to identify the earliest step at which the context is sufficient to answer the question.
We sample $K$ intermediate answers $a_i^{(k)} \sim \pi_\theta(\cdot \mid s_i, p_{\text{force}})$, where $p_{\text{force}}$ is the forced-answer instruction in Appendix~\ref{app:prompts}.
The judge scores each intermediate answer against the reference answer $a^{\text{ref}}$.
When a reference answer is unavailable, as in the expert trajectories, we use the final answer of the trajectory as $a^{\text{ref}}$.
JET therefore also applies when reference answers are unavailable.

The \emph{sufficiency} $\bar{S}_i$ of $s_i$ is the mean score of the $K$ intermediate answers,
\begin{equation}
    \bar{S}_i = \frac{1}{K} \sum_{k=1}^{K} S\bigl(a_i^{(k)}, a^{\text{ref}}\bigr) . \label{eq:sufficiency}
\end{equation}
Formally, the golden step $i^\star$ is the earliest step with the highest $\bar{S}_i$ in Eq.~\ref{eq:sufficiency},
\begin{equation}
    i^\star = \min \operatorname*{arg\,max}_{i \in \{0, \dots, n\}} \bar{S}_i . \label{eq:golden-step}
\end{equation}
The highest-scoring intermediate answer at $i^\star$ is the \emph{golden step answer} $\hat{a}$.
Every step after $i^\star$ increases the inference cost but does not improve the score.
We therefore use $i^\star$ as a proxy for the step at which the gathered information is sufficient.

\subsection{Rationale Insertion}
\label{sec:rationale}

JET truncates the trajectory at $i^\star$ and inserts a rationale $r$.
The rationale is a short message that names the files and functions examined so far and states that the gathered information is sufficient to answer.
An LLM stronger than the agent serves as the rationale generator and writes $r$ from $s_{i^\star}$ and $\hat{a}$ with the prompt in Appendix~\ref{app:prompts}.
The trajectory then ends with $y_{i^\star+1} = (r, \hat{a})$ and serves as the training data of JET.
When $i^\star = n$, the golden step is the final step of the trajectory, so we keep the original final output $y_{n+1}$.
In this output, the text preceding the answer serves as $r$, and the original final answer serves as $\hat{a}$.

\subsection{Training with a Halting Weight}
\label{sec:objective}

We define the \emph{halting span} $H$ as the tokens of the last assistant turn preceding the \texttt{<finish>} block, which contains the rationale $r$.
The agent decides to answer only in $H$, but $H$ is a small fraction of the output tokens in a trajectory.
Because the standard SFT loss is applied uniformly to all output tokens, the decision to answer contributes little to the total loss.
We therefore upweight the tokens in $H$.
The output token $x_t$ at position $t$ has a weight
\begin{equation}
    w_t =
    \begin{cases}
        \lambda & \text{if } x_t \in H, \\
        1 & \text{otherwise,}
    \end{cases}
    \label{eq:weight}
\end{equation}
where $\lambda > 1$ is the halting weight.
With the weights in Eq.~\ref{eq:weight}, we fine-tune $\pi_\theta$ on the training trajectories with a weighted cross-entropy loss over the output tokens of each trajectory,
\begin{equation}
    \mathcal{L}(\theta) = -\frac{\sum_t w_t \log \pi_\theta(x_t \mid x_{<t})}{\sum_t w_t} .
    \label{eq:loss}
\end{equation}
According to Eq.~\ref{eq:loss}, the loss is normalized by the sum of the weights, so a larger $\lambda$ shifts the loss toward $H$ without increasing the magnitude of the loss.
\section{Experiments}
\label{sec:experiments}

We conduct experiments to answer the following research questions.
\begin{itemize}[leftmargin=*,itemsep=2pt]
\item \textbf{RQ1:} Does JET reduce tool calls while preserving the score of the agent before fine-tuning?
\item \textbf{RQ2:} Does JET on expert trajectories reduce tool calls while retaining the score gain of SFT?
\item \textbf{RQ3:} How does each component of JET affect the number of tool calls and the score?
\end{itemize}
For RQ1 and RQ2, we also compare JET with existing methods for efficient tool use.

\subsubsection{Benchmarks and Agents.}
\label{sec:benchmarks}
\label{sec:models}

We evaluate JET on two repository-level QA benchmarks, SWE-QA-Pro~\citep{cai-etal-2026-swe} with 260 questions and SWE-QA~\citep{peng-etal-2026-swe} with 720 questions.
We score the answers following the judge protocol of each benchmark, in which an LLM judge compares the answer with a reference answer on five criteria, namely correctness, completeness, relevance, clarity, and reasoning.
We run both benchmarks on the SWE-QA-Pro scaffold~\citep{cai-etal-2026-swe}, which provides a substring search over the repository, a file viewer, and a read-only shell as tools.
For the agent model $\pi_\theta$, we adopt two open-weight 8B models, Qwen3-8B~\citep{qwen3technicalreport} and Ministral-3-8B-Instruct~\citep{liu2026ministral3}, and refer to each model before fine-tuning as the base agent.
Appendix~\ref{app:setup} lists the exact versions of all models.

\begin{table*}[t]
\centering
\small
\def\pct#1#2{\hspace{0.4em}{\small\textcolor[RGB]{90,90,90}{\makebox[0.8em][c]{#1}\makebox[1.9em][r]{#2\%}}}}
\def\pctblank{\hspace{0.4em}{\small\makebox[2.7em]{}}}
\renewcommand{\arraystretch}{1.15}
\begin{tabular}{l!{\color{black!35}\vrule}l!{\color{black!35}\vrule}ccc!{\color{black!35}\vrule}ccc}
\toprule
\textbf{Training data} & \textbf{Method} & \multicolumn{3}{c!{\color{black!35}\vrule}}{\textbf{SWE-QA-Pro}} & \multicolumn{3}{c}{\textbf{SWE-QA}} \\
 & & Score~$\uparrow$ & Calls~$\downarrow$ & Cost~(\$)$\downarrow$ & Score~$\uparrow$ & Calls~$\downarrow$ & Cost~(\$)$\downarrow$ \\
\midrule
\rowcolor{tabhead}\multicolumn{8}{l}{\textbf{Qwen3-8B}} \\
\multirow{5}{*}{Self} & Base & 30.88 & 18.10\pctblank & 19.2 & 36.50 & 19.57\pctblank & 23.0 \\
 & Budget Tracker & \cellcolor{tabbad!10}30.29 & \cellcolor{tabbad!44}26.12\pct{$+$}{44} & \cellcolor{tabbad!47}28.2 & \cellcolor{tabbad!22}34.88 & \cellcolor{tabbad!31}25.54\pct{$+$}{31} & \cellcolor{tabbad!21}27.8 \\
 & Autorater & \cellcolor{tabbad!3}30.70 & \cellcolor{tabgood!16}15.23\pct{$-$}{16} & \cellcolor{tabbad!50}29.0 & \cellcolor{tabbad!12}35.61 & \cellcolor{tabgood!17}16.18\pct{$-$}{17} & \cellcolor{tabbad!44}33.2 \\
 & Trunc.\ SFT  & \cellcolor{tabgood!6}31.22 & \cellcolor{tabgood!26}13.37\pct{$-$}{26} & \cellcolor{tabgood!46}10.3 & \cellcolor{tabgood!6}36.91 & \cellcolor{tabgood!31}13.54\pct{$-$}{31} & \cellcolor{tabgood!49}11.7 \\
 & \textbf{JET} & 30.84 & \cellcolor{tabgood!33}12.19\pct{$-$}{33} & \cellcolor{tabgood!50}8.5 & \cellcolor{tabbad!2}36.34 & \cellcolor{tabgood!33}13.21\pct{$-$}{33} & \cellcolor{tabgood!50}10.0 \\
\arrayrulecolor{black!35}\specialrule{0.4pt}{0pt}{0pt}\arrayrulecolor{black}
\multirow{5}{*}{Expert} & SFT & 37.07 & 22.66\pctblank & 38.3 & 52.79 & 22.18\pctblank & 36.8 \\
 & Budget Tracker & 37.08 & \cellcolor{tabbad!6}24.05\pct{$+$}{6} & \cellcolor{tabbad!14}43.7 & \cellcolor{tabgood!3}53.09 & \cellcolor{tabbad!4}23.17\pct{$+$}{4} & \cellcolor{tabbad!12}41.3 \\
 & Autorater & \cellcolor{tabbad!35}34.49 & \cellcolor{tabgood!50}11.31\pct{$-$}{50} & \cellcolor{tabgood!29}27.4 & \cellcolor{tabbad!41}48.42 & \cellcolor{tabgood!50}10.64\pct{$-$}{52} & \cellcolor{tabgood!31}25.5 \\
 & Trunc.\ SFT  & \cellcolor{tabbad!7}36.52 & \cellcolor{tabgood!3}21.90\pct{$-$}{3} & \cellcolor{tabgood!3}37.2 & \cellcolor{tabbad!9}51.83 & \cellcolor{tabgood!4}21.35\pct{$-$}{4} & \cellcolor{tabgood!5}35.0 \\
 & \textbf{JET} & \cellcolor{tabbad!15}35.93 & \cellcolor{tabgood!26}16.72\pct{$-$}{26} & \cellcolor{tabgood!36}24.4 & \cellcolor{tabbad!50}47.47 & \cellcolor{tabgood!28}15.90\pct{$-$}{28} & \cellcolor{tabgood!39}22.4 \\
\arrayrulecolor{black!35}\specialrule{0.4pt}{0pt}{0pt}\arrayrulecolor{black}
\multirow{2}{*}{Expert $\rightarrow$ Self} & \textbf{JET} $\rightarrow$ Trunc.\ SFT  & \cellcolor{tabbad!24}35.29 & \cellcolor{tabgood!46}12.17\pct{$-$}{46} & \cellcolor{tabgood!50}14.9 & \cellcolor{tabbad!50}46.59 & \cellcolor{tabgood!50}10.85\pct{$-$}{51} & \cellcolor{tabgood!50}12.3 \\
 & \textbf{JET} $\rightarrow$ \textbf{JET} & \cellcolor{tabbad!27}35.08 & \cellcolor{tabgood!43}12.87\pct{$-$}{43} & \cellcolor{tabgood!50}16.5 & \cellcolor{tabbad!50}46.25 & \cellcolor{tabgood!44}12.40\pct{$-$}{44} & \cellcolor{tabgood!50}15.4 \\
\midrule
\rowcolor{tabhead}\multicolumn{8}{l}{\textbf{Ministral-3-8B}} \\
\multirow{5}{*}{Self} & Base & 35.84 & 14.71\pctblank & 17.8 & 46.27 & 14.68\pctblank & 19.1 \\
 & Budget Tracker & 35.75 & 14.88\pct{$+$}{1} & \cellcolor{tabbad!11}19.7 & 46.38 & 14.86\pct{$+$}{1} & \cellcolor{tabbad!8}20.7 \\
 & Autorater & \cellcolor{tabbad!26}33.96 & \cellcolor{tabgood!26}10.82\pct{$-$}{26} & \cellcolor{tabbad!24}22.0 & \cellcolor{tabbad!4}45.93 & \cellcolor{tabgood!26}10.79\pct{$-$}{26} & \cellcolor{tabbad!18}22.5 \\
 & Trunc.\ SFT  & \cellcolor{tabbad!4}35.53 & \cellcolor{tabbad!2}14.96\pct{$+$}{2} & \cellcolor{tabgood!3}17.3 & 46.32 & \cellcolor{tabgood!4}14.03\pct{$-$}{4} & \cellcolor{tabgood!12}16.9 \\
 & \textbf{JET} & \cellcolor{tabbad!14}34.81 & \cellcolor{tabgood!12}12.94\pct{$-$}{12} & \cellcolor{tabgood!19}14.4 & \cellcolor{tabbad!3}45.97 & \cellcolor{tabgood!11}13.10\pct{$-$}{11} & \cellcolor{tabgood!25}14.3 \\
\arrayrulecolor{black!35}\specialrule{0.4pt}{0pt}{0pt}\arrayrulecolor{black}
\multirow{5}{*}{Expert} & SFT & 35.10 & 23.74\pctblank & 42.1 & 51.78 & 22.37\pctblank & 39.0 \\
 & Budget Tracker & \cellcolor{tabgood!12}35.94 & \cellcolor{tabgood!2}23.37\pct{$-$}{2} & \cellcolor{tabbad!2}42.8 & \cellcolor{tabgood!6}52.37 & 22.24\pct{$-$}{1} & \cellcolor{tabbad!6}41.3 \\
 & Autorater & \cellcolor{tabbad!9}34.48 & \cellcolor{tabgood!43}13.56\pct{$-$}{43} & \cellcolor{tabgood!24}31.9 & \cellcolor{tabbad!11}50.64 & \cellcolor{tabgood!45}12.21\pct{$-$}{45} & \cellcolor{tabgood!29}27.9 \\
 & Trunc.\ SFT  & \cellcolor{tabgood!27}37.02 & \cellcolor{tabbad!10}26.08\pct{$+$}{10} & \cellcolor{tabbad!12}47.2 & \cellcolor{tabgood!24}54.22 & \cellcolor{tabbad!15}25.65\pct{$+$}{15} & \cellcolor{tabbad!16}45.4 \\
 & \textbf{JET} & \cellcolor{tabgood!26}36.91 & \cellcolor{tabgood!20}19.10\pct{$-$}{20} & \cellcolor{tabgood!27}30.6 & \cellcolor{tabgood!20}53.83 & \cellcolor{tabgood!18}18.36\pct{$-$}{18} & \cellcolor{tabgood!27}28.6 \\
\arrayrulecolor{black!35}\specialrule{0.4pt}{0pt}{0pt}\arrayrulecolor{black}
\multirow{2}{*}{Expert $\rightarrow$ Self} & \textbf{JET} $\rightarrow$ Trunc.\ SFT  & \cellcolor{tabgood!8}35.67 & \cellcolor{tabgood!17}19.75\pct{$-$}{17} & \cellcolor{tabgood!22}32.9 & \cellcolor{tabbad!12}50.49 & \cellcolor{tabgood!18}18.43\pct{$-$}{18} & \cellcolor{tabgood!24}29.7 \\
 & \textbf{JET} $\rightarrow$ \textbf{JET} & \cellcolor{tabbad!7}34.63 & \cellcolor{tabgood!45}12.95\pct{$-$}{45} & \cellcolor{tabgood!50}15.9 & \cellcolor{tabbad!31}48.55 & \cellcolor{tabgood!43}12.84\pct{$-$}{43} & \cellcolor{tabgood!50}16.5 \\
\bottomrule
\end{tabular}
\caption{Main results on SWE-QA-Pro and SWE-QA. Cost is in US dollars per 1,000 questions. Purple shading marks an improvement and red shading a degradation, with darker shading for a larger difference. Shading and percentages are relative to Base in the Self setting and to SFT in the other two settings. Trunc.\ SFT denotes Truncated SFT (\S\ref{sec:baselines}).}
\label{tab:main-results}
\end{table*}

\subsection{Training}
\label{sec:training}

\paragraph{Data synthesis.}
We apply JET to two types of trajectories, self-generated trajectories (\textbf{Self}) and expert trajectories (\textbf{Expert}).
Self trajectories are the rollouts of the base agent on 400 training questions from SWE-QA-Pro~\citep{cai-etal-2026-swe}, which do not overlap with the benchmark questions.
Each question is paired with a reference answer, which serves as $a^{\text{ref}}$.
Expert trajectories are the 1,000 SFT trajectories released with SWE-QA-Pro, generated by Claude Sonnet 4.5.
In Expert $\rightarrow$ Self, the agent trained with JET in the Expert setting generates rollouts on the 400 training questions of the Self setting, and we fine-tune the agent again on the rollouts.

We hold out 60 additional training questions as the validation set for model selection.
In golden step labeling, we sample $K = 3$ intermediate answers at every step with temperature 0.6, and GPT-6 Luna scores the answers.
In rationale insertion, Qwen3.8-27B~\citep{qwen38} serves as the rationale generator.

\paragraph{Fine-tuning.}
We set the halting weight to $\lambda = 5$ for all runs, selected on the validation set with Ministral in the Self setting.
All models are fine-tuned with LoRA~\citep{hu2022lora} on NVIDIA RTX PRO 6000 and H100 GPUs.
We use the same hyperparameters for all methods, with a rank of 32, a learning rate of $5 \times 10^{-4}$, and an effective batch size of 16 trajectories.
We train for up to 4 epochs.
Appendix~\ref{app:setup} lists all training and evaluation hyperparameters.

\subsection{Baselines}
\label{sec:baselines}

We compare JET with four baselines, which are SFT, Truncated SFT, Budget Tracker, and the autorater.
In the Self setting, the comparison is against the base agent, whose own rollouts serve as the training trajectories (RQ1).
In the Expert and Expert $\rightarrow$ Self settings, the comparison is against SFT, which is trained on the same expert trajectories (RQ2).
Truncated SFT examines the effect of the rationale and the halting weight (RQ3).
Budget Tracker and the autorater are training-free methods, and we apply both to the base and SFT agents.

\paragraph{SFT.}
SFT follows the supervised fine-tuning stage of SWE-QA-Pro~\citep{cai-etal-2026-swe}, which fine-tunes the agent on the 1,000 expert trajectories to improve task performance.
SFT shows the effect of the expert trajectories on the score and tool calls when efficiency is not considered.

\paragraph{Truncated SFT.}
Truncated SFT is standard SFT on the trajectories truncated at the golden step, without the rationale and the halting weight.
Truncated SFT shows whether truncation alone is sufficient to reduce tool calls.

\paragraph{Budget Tracker.}
Budget Tracker~\citep{liu2026budgetaware} states a budget of tool calls in the prompt and appends the used and remaining budget after every tool call, together with a guideline for using the budget.
We set the budget to 29 tool calls and adapt the guideline to our agent, as described in Appendix~\ref{app:prompts}.
Budget Tracker shows whether a prompting method can reduce tool calls without training.

\paragraph{Autorater.}
The autorater follows the sufficient-context autorater of \citet{ICLR2025_33dffa2e} and asks a frontier LLM whether the context is sufficient to answer the question.
After every step, GPT-6 Luna reads the question and the current context.
Once the autorater labels the context as sufficient, we force the agent to answer.
If the agent does not produce an answer, the agent proceeds to the next step.
We adapt the prompt to our agent, as described in Appendix~\ref{app:prompts}.
The autorater shows whether a stronger LLM can decide when to answer in place of the agent.

\subsection{Evaluation}
\label{sec:evaluation}

We run all agents with temperature 0, a context window of 32K tokens, and a step limit of 30, at which the scaffold forces the agent to answer.

\paragraph{Checkpoint selection.}
We save a checkpoint after every epoch and evaluate each checkpoint on the validation set.
For SFT, we select the checkpoint with the highest score.
For JET and Truncated SFT, we select the checkpoint with the highest score among the checkpoints that make fewer tool calls than the base agent in the Self setting and than SFT in the other settings.

\paragraph{Metrics.}
\label{sec:metrics}
\textbf{Score} is the judge score of the final answer averaged over the questions, with a maximum of 50 on SWE-QA-Pro and 100 on SWE-QA.
The judges are GPT-6 Sol on SWE-QA-Pro and Claude Sonnet 5.5 on SWE-QA, which are newer versions of the judges in the original papers.
\textbf{Tool calls} is the average number of tool calls per question before the final answer.
\textbf{Cost} is the inference cost per 1,000 questions in US dollars, including the LLM calls of the autorater.
Appendix~\ref{app:eval} describes the pricing and how we score the questions on which the agent fails to answer.

\subsection{Main Results}
\label{sec:main-results}

Table~\ref{tab:main-results} presents the main results on both benchmarks.
JET is the only method that reduces both tool calls and inference cost in every setting, without any extra LLM call at inference time.

\subsubsection{Comparison with the Base Agent (RQ1).}
JET reduces tool calls while largely preserving the score.
In the Self setting, JET reduces tool calls by 11 to 33\% on every model and benchmark, and the score stays within about one point of the base agent.
The additional tool calls of the base agent are therefore unnecessary, which indicates that JET mitigates over-exploration.
JET also has the lowest cost among all methods in the Self setting.
In contrast, no baseline reduces both tool calls and cost on both models.
Budget Tracker does not reduce tool calls on either model and raises the cost on both benchmarks.
With Budget Tracker, the tool calls of Ministral remain nearly unchanged, whereas Qwen3 makes up to 44\% more tool calls, because Qwen3 follows the guideline to explore broadly.
This result indicates that the effect of prompting depends on how each model follows the instruction and does not generalize across models.
The autorater makes fewer tool calls than the base agent, but degrades the score more than JET and costs more than the base agent, because the autorater calls a frontier LLM at every step.
JET does not incur the additional cost, since the agent itself decides when to answer.
Truncated SFT lowers the tool calls of Qwen3 by 26 to 31\%, but changes the tool calls of Ministral by at most 4\%, which indicates that the effect of truncation alone depends on the model. 

\subsubsection{Comparison with SFT (RQ2).}
JET retains most of the score gain of SFT with fewer tool calls and a lower cost.
SFT on expert trajectories improves the score in three of the four cases, but also substantially increases tool calls over the base agent.
JET on the same trajectories makes 18 to 28\% fewer tool calls than SFT on every model and benchmark, at a lower cost.
On Qwen3, JET retains most of the score gain of SFT and makes fewer tool calls than even the base agent.
On Ministral, JET outperforms SFT in score, tool calls, and cost, which we analyze in \S\ref{sec:stopping}.
In contrast, the baselines either change tool calls only slightly or lose more of the score.
Budget Tracker has almost no effect on the tool calls of SFT.
Truncated SFT, which truncates the trajectories without the rationale and the halting weight, lowers tool calls only on Qwen3 and only marginally, and increases tool calls on Ministral.
The effect of truncation alone is inconsistent across models, so the rationale and the halting weight are crucial components for reducing tool calls on both models.
The~autorater makes fewer tool calls than JET, but scores lower or costs more than JET in all four cases.

\subsubsection{Two-Stage Training (Expert $\rightarrow$ Self).}
In the Expert setting, the golden steps are labeled on the expert trajectories, not on the agent's own rollouts.
We further examine a second stage of training in the Expert $\rightarrow$ Self setting, which applies JET once more to the rollouts of the trained agent.
The second stage reduces tool calls further.
The resulting agent makes 43 to 45\% fewer tool calls than SFT in all four cases, at less than half of the cost of SFT.
The resulting agent also makes fewer tool calls than the base agent, at a lower cost.
Compared with JET in the Expert setting, tool calls decrease by 22 to 32\%.
A second stage of Truncated SFT achieves a comparable reduction on Qwen3, while the reduction on Ministral is less pronounced.
The score remains above the base agent in three of the four cases.
Compared with JET in the Expert setting, the score decreases by 0.9 to 5.3 points.
The second stage therefore trades part of the score for further efficiency.
\section{Analysis}
\label{sec:analysis}

\begin{figure}[!t]
\centering
\includegraphics[width=\columnwidth]{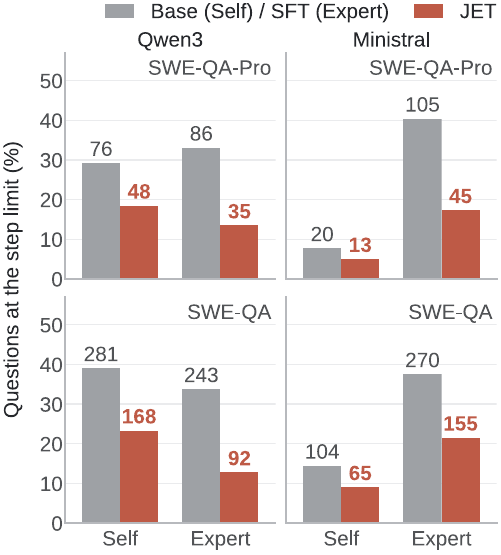}
\caption{Percentage of questions on which the agent reaches the step limit. The numbers above the bars are the question counts. JET is compared with Base in the Self setting and with SFT in the Expert setting.}
\label{fig:stepLimit}
\end{figure}

\begin{table}[t]
\centering
\small
\def\pct#1#2{\hspace{0.4em}{\small\textcolor[RGB]{90,90,90}{\makebox[0.8em][c]{#1}\makebox[1.9em][r]{#2\%}}}}
\def\pctblank{\hspace{0.4em}{\small\makebox[2.7em]{}}}
\renewcommand{\arraystretch}{1.15}
\begin{tabular}{llcc}
\toprule
\multicolumn{2}{l}{\textbf{Truncation step}} & Score~$\uparrow$ & Calls~$\downarrow$ \\
\midrule
\rowcolor{tabhead}\multicolumn{4}{l}{\textbf{Qwen3-8B}} \\
 & Base & 30.88 & 18.10\pctblank \\
\arrayrulecolor{black!35}\specialrule{0.4pt}{0pt}{0pt}\arrayrulecolor{black}
\multirow{3}{*}{Truncation} & $n/2$ & \cellcolor{tabbad!19}29.71 & \cellcolor{tabgood!12}15.90\pct{$-$}{12} \\
 & $2n/3$ & 30.91 & \cellcolor{tabbad!16}20.94\pct{$+$}{16} \\
 & $i^\star$ (\textbf{JET}) & 30.84 & \cellcolor{tabgood!33}12.19\pct{$-$}{33} \\
\midrule
\rowcolor{tabhead}\multicolumn{4}{l}{\textbf{Ministral-3-8B}} \\
 & Base & 35.84 & 14.71\pctblank \\
\arrayrulecolor{black!35}\specialrule{0.4pt}{0pt}{0pt}\arrayrulecolor{black}
\multirow{3}{*}{Truncation} & $n/2$ & \cellcolor{tabbad!50}29.68 & \cellcolor{tabgood!50}6.82\pct{$-$}{54} \\
 & $2n/3$ & \cellcolor{tabbad!50}31.95 & \cellcolor{tabgood!39}8.93\pct{$-$}{39} \\
 & $i^\star$ (\textbf{JET}) & \cellcolor{tabbad!14}34.81 & \cellcolor{tabgood!12}12.94\pct{$-$}{12} \\
\bottomrule
\end{tabular}
\caption{Ablation of the golden step on SWE-QA-Pro in the Self setting. The rows $n/2$ and $2n/3$ truncate each trajectory at a fixed ratio of the number of steps $n$. Shading and percentages are relative to Base.}
\label{tab:golden-step}
\end{table}

\begin{figure}[!t]
\centering
\includegraphics[width=\linewidth]{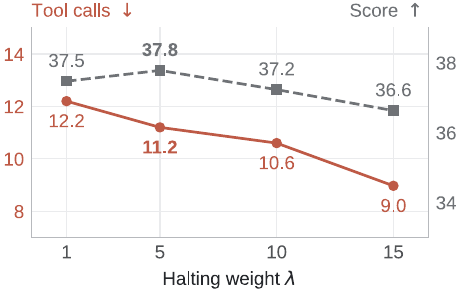}
\caption{Tool calls and score of Ministral in the Self setting on the validation set for each halting weight $\lambda$.}
\label{fig:lambda}
\end{figure}
\begin{figure*}[t]
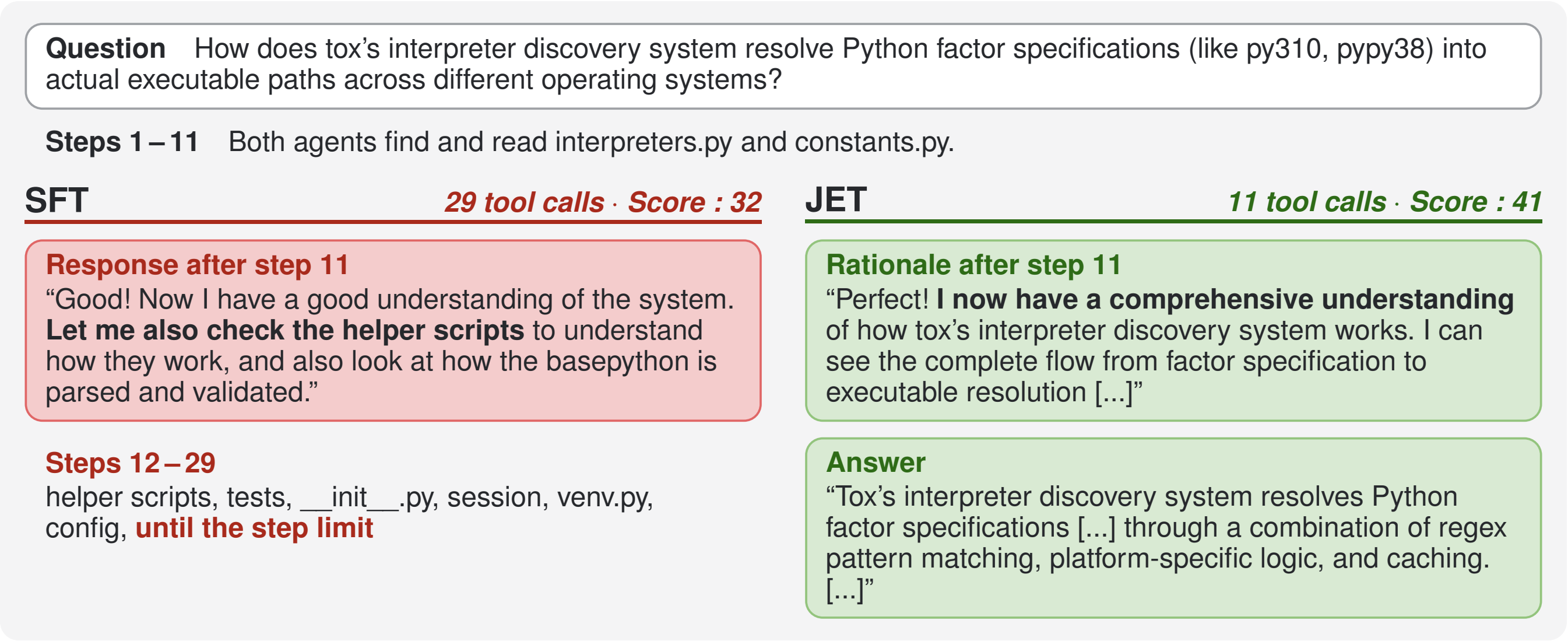

\definecolor{cspanel}{HTML}{F4F4F5}
\definecolor{csink}{HTML}{26282C}
\definecolor{csline}{HTML}{9EA1A5}
\definecolor{csgreenbg}{HTML}{D9EAD3}
\definecolor{csgreenln}{HTML}{93C47D}
\definecolor{csgreen}{HTML}{2E6B17}
\definecolor{csredbg}{HTML}{F4CCCC}
\definecolor{csredln}{HTML}{E06666}
\definecolor{csred}{HTML}{A82A1C}
\tcbset{
  csbubble/.style={enhanced, boxrule=0.7pt, arc=5pt, left=6pt, right=6pt, top=4pt, bottom=4pt, boxsep=0pt,
    fontupper=\small\sffamily\color{csink}, halign=flush left},
  csplain/.style={enhanced, frame hidden, interior hidden, boxrule=0pt, boxsep=0pt, left=6.7pt, right=6.7pt, top=4.7pt, bottom=4.7pt, valign=top,
    fontupper=\small\sffamily\color{csink}, halign=flush left},
  cshead/.style={blanker, fontupper=\small\sffamily\color{csink}}}
\newcommand{\cshead}[3]{{\large\bfseries #1}\hfill \textcolor{#2}{\textbf{\textit{#3}}}\par\vspace{2.5pt}{\color{#2}\hrule height 1.2pt}}
\begin{tcolorbox}[enhanced, colback=cspanel, frame hidden, boxrule=0pt, arc=6pt,
  left=8pt, right=8pt, top=7pt, bottom=7pt, boxsep=0pt, before skip=0pt, after skip=4pt]
\begin{tcolorbox}[csbubble, colback=white, colframe=csline, before skip=0pt, after skip=2pt]
\textbf{Question}\quad How does tox's interpreter discovery system resolve Python factor specifications (like py310, pypy38) into actual executable paths across different operating systems?
\end{tcolorbox}
\begin{tcolorbox}[csplain, before skip=0pt, after skip=5pt]
\textbf{Steps 1\,--\,11}\quad Both agents find and read interpreters.py and constants.py.
\end{tcolorbox}
\begin{tcbraster}[raster columns=2, raster equal height=rows, raster column skip=14pt, raster row skip=5pt,
  raster before skip=0pt, raster after skip=0pt]
\begin{tcolorbox}[cshead]
\cshead{SFT}{csred}{29 tool calls $\cdot$ Score : 32}
\end{tcolorbox}
\begin{tcolorbox}[cshead]
\cshead{JET}{csgreen}{11 tool calls $\cdot$ Score : 41}
\end{tcolorbox}
\begin{tcolorbox}[csbubble, colback=csredbg, colframe=csredln]
\textcolor{csred}{\textbf{Response after step 11}}\par\vspace{1pt}
``Good! Now I have a good understanding of the system. \textbf{Let me also check the helper scripts} to understand how they work, and also look at how the basepython is parsed and validated.''
\end{tcolorbox}
\begin{tcolorbox}[csbubble, colback=csgreenbg, colframe=csgreenln]
\textcolor{csgreen}{\textbf{Rationale after step 11}}\par\vspace{1pt}
``Perfect! \textbf{I now have a comprehensive understanding} of how tox's interpreter discovery system works. I can see the complete flow from factor specification to executable resolution [...]''
\end{tcolorbox}
\begin{tcolorbox}[csplain]
\textcolor{csred}{\textbf{Steps 12\,--\,29}}\par\vspace{1pt}
helper scripts, tests, \_\_init\_\_.py, session, venv.py, config, \textcolor{csred}{\textbf{until the step limit}}
\end{tcolorbox}
\begin{tcolorbox}[csbubble, colback=csgreenbg, colframe=csgreenln]
\textcolor{csgreen}{\textbf{Answer}}\par\vspace{1pt}
``Tox's interpreter discovery system resolves Python factor specifications [...] through a combination of regex pattern matching, platform-specific logic, and caching. [...]''
\end{tcolorbox}
\end{tcbraster}
\end{tcolorbox}
\caption{Case study with Qwen3 in the Expert setting on a question from SWE-QA-Pro. Both agents read the relevant code within 11 tool calls. SFT continues to invoke tools until the step limit, whereas JET answers after step 11. Scores range from 5 to 50. We summarize the tool calls and shorten the responses, marked by [...].}
\label{fig:case-study}
\end{figure*}

\subsection{Answering Behavior (RQ1 \& RQ2)}
\label{sec:stopping}
To examine whether JET trains the agent to answer before being forced, we measure how often each agent reaches the step limit, at which the scaffold forces the agent to answer (Figure~\ref{fig:stepLimit}).
JET reaches the step limit less often on every model and benchmark, by 43 to 62\% relative to SFT.
With Qwen3 on SWE-QA-Pro, JET answers before the limit on 64 of the 86 questions on which SFT reaches the step limit.
Consequently, JET makes fewer tool calls by answering before the step limit.
In a case study (Figure~\ref{fig:case-study}), both agents read the relevant code in two files within 11 tool calls.
SFT then makes 18 more tool calls until the step limit, whereas JET answers after reading the code.
As a result, JET achieves a higher score with less than half of the tool calls.

The step limit also explains why JET scores higher than SFT on Ministral.
On Ministral, the SFT agent reaches the step limit on about 40\% of the questions, far more often than the base agent.
On 6 to 7\% of the questions, the SFT agent does not produce a valid answer even when forced, and the judge assigns a score near the minimum of the scale.
The base agent and JET in the Expert setting produce a valid answer on almost all questions.

\subsection{Ablation Study (RQ3)}
\label{sec:ablation}

We conduct an ablation study to examine how each component of JET affects tool calls and the score.

\paragraph{Golden step.}
We replace the golden step with a fixed truncation step, $n/2$ or $2n/3$, and report the results in Table~\ref{tab:golden-step}.
On Qwen3, a fixed truncation step reduces tool calls less than the golden step or even increases tool calls.
On Ministral, a fixed truncation step makes fewer tool calls than the golden step, but lowers the score substantially.
The step at which the gathered information is sufficient differs across questions, so a fixed ratio truncates some trajectories before the information is sufficient and others long after.
This result shows the need for the golden step.

\paragraph{Rationale and halting weight.}
The rationale is effective only together with the halting weight.
Truncated SFT omits both components.
Truncated SFT reduces tool calls less than JET in the Self and Expert settings, and even increases tool calls on Ministral in the Expert setting (Table~\ref{tab:main-results}).
We then vary $\lambda$ on the validation set with Ministral in the Self setting to examine the effect of the halting weight (Figure~\ref{fig:lambda}).
With $\lambda = 1$, the agent is trained with the rationale alone.
The agent makes 12.20 tool calls, close to Truncated SFT.
As $\lambda$ increases, tool calls decrease further.
The halting weight therefore controls the trade-off between tool calls and the score.
\section{Conclusion}
In this paper, we introduced JET, a fine-tuning method that mitigates over-exploration in repository-level QA by training agents to answer once the gathered information is sufficient.
We found that over-exploration appears in both agents and the expert trajectories for SFT, where the score reaches a plateau while the exploration continues.
Across two models and two benchmarks, JET is the only method that reduces both tool calls and inference cost in every setting.
On expert trajectories, JET retains most of the score gain of SFT with substantially fewer tool calls, and even scores higher than SFT with Ministral.
These results show that an agent can use tools efficiently while largely preserving task performance.
We hope that this work encourages future research on agents to consider efficient tool use as well as task performance.
\section*{Limitations}
Our work has four limitations.
First, the experiments are limited to the code domain.
Both benchmarks consist of questions about Python repositories, and we evaluate two 8B models on one scaffold.
All results come from a single training run for each setting.
Whether JET generalizes to larger models, repositories in other programming languages, and other long-horizon tasks with free-form answers remains to be verified.

Second, the golden step is labeled by an LLM judge from $K = 3$ sampled answers per step, and we do not verify the labels with human annotators.
The labels are therefore subject to the variance of the judge.

Third, we evaluate the trained agents by the score, the number of tool calls, and how often each agent reaches the step limit.
We do not identify which tool calls JET removes, and leave a step-level analysis of the trained agents to future work.

Fourth, golden step labeling samples $K$ intermediate answers at every step of every training trajectory and scores each answer with the judge.
This cost is paid once and offline, but grows with the number and length of the training trajectories.
For this reason, we do not apply JET to the rollouts of the SFT agents, and leave this setting to future work.
\bibliography{custom}

@misc{chen2021evaluatinglargelanguagemodels,
      title={Evaluating Large Language Models Trained on Code}, 
      author={Mark Chen and Jerry Tworek and Heewoo Jun and Qiming Yuan and Henrique Ponde de Oliveira Pinto and Jared Kaplan and Harri Edwards and Yuri Burda and Nicholas Joseph and Greg Brockman and Alex Ray and Raul Puri and Gretchen Krueger and Michael Petrov and Heidy Khlaaf and Girish Sastry and Pamela Mishkin and Brooke Chan and Scott Gray and Nick Ryder and Mikhail Pavlov and Alethea Power and Lukasz Kaiser and Mohammad Bavarian and Clemens Winter and Philippe Tillet and Felipe Petroski Such and Dave Cummings and Matthias Plappert and Fotios Chantzis and Elizabeth Barnes and Ariel Herbert-Voss and William Hebgen Guss and Alex Nichol and Alex Paino and Nikolas Tezak and Jie Tang and Igor Babuschkin and Suchir Balaji and Shantanu Jain and William Saunders and Christopher Hesse and Andrew N. Carr and Jan Leike and Josh Achiam and Vedant Misra and Evan Morikawa and Alec Radford and Matthew Knight and Miles Brundage and Mira Murati and Katie Mayer and Peter Welinder and Bob McGrew and Dario Amodei and Sam McCandlish and Ilya Sutskever and Wojciech Zaremba},
      year={2021},
      eprint={2107.03374},
      archivePrefix={arXiv},
      primaryClass={cs.LG},
      url={https://arxiv.org/abs/2107.03374}, 
}

@inproceedings{
yao2023react,
title={ReAct: Synergizing Reasoning and Acting in Language Models},
author={Shunyu Yao and Jeffrey Zhao and Dian Yu and Nan Du and Izhak Shafran and Karthik R Narasimhan and Yuan Cao},
booktitle={The Eleventh International Conference on Learning Representations },
year={2023},
url={https://openreview.net/forum?id=WE_vluYUL-X}
}

@inproceedings{
    jimenez2024swebench,
    title={{SWE}-bench: Can Language Models Resolve Real-world Github Issues?},
    author={Carlos E Jimenez and John Yang and Alexander Wettig and Shunyu Yao and Kexin Pei and Ofir Press and Karthik R Narasimhan},
    booktitle={The Twelfth International Conference on Learning Representations},
    year={2024},
    url={https://openreview.net/forum?id=VTF8yNQM66}
}

@inproceedings{yang2024sweagent,
  title={{SWE}-agent: Agent-Computer Interfaces Enable Automated Software Engineering},
  author={John Yang and Carlos E Jimenez and Alexander Wettig and Kilian Lieret and Shunyu Yao and Karthik R Narasimhan and Ofir Press},
  booktitle={The Thirty-eighth Annual Conference on Neural Information Processing Systems},
  year={2024},
  url={https://arxiv.org/abs/2405.15793}
}

@inproceedings{peng-etal-2026-swe,
    title = "{SWE}-{QA}: Can Language Models Answer Repository-level Code Questions?",
    author = "Peng, Weihan  and
      Shi, Yuling  and
      Wang, Yuhang  and
      Zhang, Xinyun  and
      Shen, Beijun  and
      Gu, Xiaodong",
    editor = "Liakata, Maria  and
      Moreira, Viviane P.  and
      Zhang, Jiajun  and
      Jurgens, David",
    booktitle = "Findings of the {A}ssociation for {C}omputational {L}inguistics: {ACL} 2026",
    month = jul,
    year = "2026",
    address = "San Diego, California, United States",
    publisher = "Association for Computational Linguistics",
    url = "https://aclanthology.org/2026.findings-acl.402/",
    doi = "10.18653/v1/2026.findings-acl.402",
    pages = "8230--8245",
    ISBN = "979-8-89176-395-1"
}

@inproceedings{cai-etal-2026-swe,
    title = "{SWE}-{QA}-Pro: A Representative Benchmark and Scalable Training Recipe for Repository-Level Code Understanding",
    author = "Cai, Songcheng  and
      Lyu, Zhiheng  and
      Ni, Yuansheng  and
      Chen, Xiangchao  and
      Zhou, Baichuan  and
      Zhu, Shenzhe  and
      Lu, Yi  and
      Wang, Haozhe  and
      Ruan, Chi  and
      Schneider, Benjamin  and
      Zhang, Weixu  and
      Li, Xiang  and
      Zheng, Andy  and
      Zhang, Yuyu  and
      Nie, Ping  and
      Chen, Wenhu",
    editor = "Liakata, Maria  and
      Moreira, Viviane P.  and
      Zhang, Jiajun  and
      Jurgens, David",
    booktitle = "Findings of the {A}ssociation for {C}omputational {L}inguistics: {ACL} 2026",
    month = jul,
    year = "2026",
    address = "San Diego, California, United States",
    publisher = "Association for Computational Linguistics",
    url = "https://aclanthology.org/2026.findings-acl.837/",
    doi = "10.18653/v1/2026.findings-acl.837",
    pages = "16958--16999",
    ISBN = "979-8-89176-395-1"
}

@inproceedings{zeng-etal-2024-agenttuning,
    title = "{A}gent{T}uning: Enabling Generalized Agent Abilities for {LLM}s",
    author = "Zeng, Aohan  and
      Liu, Mingdao  and
      Lu, Rui  and
      Wang, Bowen  and
      Liu, Xiao  and
      Dong, Yuxiao  and
      Tang, Jie",
    editor = "Ku, Lun-Wei  and
      Martins, Andre  and
      Srikumar, Vivek",
    booktitle = "Findings of the {A}ssociation for {C}omputational {L}inguistics: {ACL} 2024",
    month = aug,
    year = "2024",
    address = "Bangkok, Thailand",
    publisher = "Association for Computational Linguistics",
    url = "https://aclanthology.org/2024.findings-acl.181/",
    doi = "10.18653/v1/2024.findings-acl.181",
    pages = "3053--3077"
}

@inproceedings{
pan2025training,
title={Training Software Engineering Agents and Verifiers with {SWE}-Gym},
author={Jiayi Pan and Xingyao Wang and Graham Neubig and Navdeep Jaitly and Heng Ji and Alane Suhr and Yizhe Zhang},
booktitle={Forty-second International Conference on Machine Learning},
year={2025},
url={https://openreview.net/forum?id=Cq1BNvHx74}
}

@inproceedings{
liu2026budgetaware,
title={Budget-Aware Tool Use Enables Effective Agent Scaling},
author={Tengxiao Liu and Zifeng Wang and Jin Miao and I-Hung Hsu and Jun Yan and Jiefeng Chen and Rujun Han and Fangyuan Xu and Yanfei Chen and Ke Jiang and Samira Daruki and Yi Liang and William Yang Wang and Tomas Pfister and Chen-Yu Lee},
booktitle={Third Conference on Language Modeling},
year={2026},
url={https://openreview.net/forum?id=hZ5XPW7jaD}
}

@inproceedings{ICLR2025_33dffa2e,
 author = {Joren, Hailey and Zhang, Jianyi and Ferng, Chun-Sung and Juan, Da-Cheng and Taly, Ankur and Rashtchian, Cyrus},
 booktitle = {International Conference on Learning Representations},
 editor = {Y. Yue and A. Garg and N. Peng and F. Sha and R. Yu},
 pages = {20310--20334},
 title = {Sufficient Context: A New Lens on Retrieval Augmented Generation Systems},
 url = {https://proceedings.iclr.cc/paper_files/paper/2025/file/33dffa2e3d2ab74a783d1a8c292f66d9-Paper-Conference.pdf},
 volume = {2025},
 year = {2025}
}

@inproceedings{sun-etal-2026-autosearch,
    title = "{A}uto{S}earch: Adaptive Search Depth for Efficient Agentic {RAG} via Reinforcement Learning",
    author = "Sun, Jingbo  and
      Chong, Wenyue  and
      Tu, Songjun  and
      Zhang, Qichao  and
      Zhang, Yaocheng  and
      Chai, Jiajun  and
      Wang, Xiaohan  and
      Lin, Wei  and
      Yin, Guojun  and
      Zhao, Dongbin",
    editor = "Liakata, Maria  and
      Moreira, Viviane P.  and
      Zhang, Jiajun  and
      Jurgens, David",
    booktitle = "Findings of the {A}ssociation for {C}omputational {L}inguistics: {ACL} 2026",
    month = jul,
    year = "2026",
    address = "San Diego, California, United States",
    publisher = "Association for Computational Linguistics",
    url = "https://aclanthology.org/2026.findings-acl.1399/",
    doi = "10.18653/v1/2026.findings-acl.1399",
    pages = "28059--28079",
    ISBN = "979-8-89176-395-1"
}

@inproceedings{10.1145/3774904.3792235,
author = {Zhang, Wenlin and Dong, Kuicai and Li, Junyi and Zhang, Yingyi and Li, Xiaopeng and Jia, Pengyue and Wen, Yi and Xu, Derong and Wang, Maolin and Wang, Yichao and Liu, Yong and Zhao, Xiangyu},
title = {To Search or Not to Search: Aligning the Decision Boundary of Deep Search Agents via Causal Intervention},
year = {2026},
isbn = {9798400723070},
publisher = {Association for Computing Machinery},
address = {New York, NY, USA},
url = {https://doi.org/10.1145/3774904.3792235},
doi = {10.1145/3774904.3792235},
booktitle = {Proceedings of the ACM Web Conference 2026},
pages = {2049–2059},
numpages = {11},
location = {Dubai, United Arab Emirates},
series = {WWW '26}
}

@misc{wang2025actingreasoningmoreteaching,
      title={Acting Less is Reasoning More! Teaching Model to Act Efficiently}, 
      author={Hongru Wang and Cheng Qian and Wanjun Zhong and Xiusi Chen and Jiahao Qiu and Shijue Huang and Bowen Jin and Mengdi Wang and Kam-Fai Wong and Heng Ji},
      year={2025},
      eprint={2504.14870},
      archivePrefix={arXiv},
      primaryClass={cs.AI},
      url={https://arxiv.org/abs/2504.14870}, 
}

@inproceedings{chen-etal-2026-et,
    title = "{ET}-Agent: Incentivizing Effective Tool-Integrated Reasoning Agent via Behavior Calibration",
    author = "Chen, Yifei  and
      Dong, Guanting  and
      Dou, Zhicheng",
    editor = "Liakata, Maria  and
      Moreira, Viviane P.  and
      Zhang, Jiajun  and
      Jurgens, David",
    booktitle = "Proceedings of the 64th Annual Meeting of the {A}ssociation for {C}omputational {L}inguistics (Volume 1: Long Papers)",
    month = jul,
    year = "2026",
    address = "San Diego, California, United States",
    publisher = "Association for Computational Linguistics",
    url = "https://aclanthology.org/2026.acl-long.333/",
    doi = "10.18653/v1/2026.acl-long.333",
    pages = "7337--7359",
    ISBN = "979-8-89176-390-6"
}

@inproceedings{qian-etal-2025-smart,
    title = "{SMART}: Self-Aware Agent for Tool Overuse Mitigation",
    author = {Qian, Cheng  and
      Acikgoz, Emre Can  and
      Wang, Hongru  and
      Chen, Xiusi  and
      Sil, Avirup  and
      Hakkani-T{\"u}r, Dilek  and
      Tur, Gokhan  and
      Ji, Heng},
    editor = "Che, Wanxiang  and
      Nabende, Joyce  and
      Shutova, Ekaterina  and
      Pilehvar, Mohammad Taher",
    booktitle = "Findings of the {A}ssociation for {C}omputational {L}inguistics: {ACL} 2025",
    month = jul,
    year = "2025",
    address = "Vienna, Austria",
    publisher = "Association for Computational Linguistics",
    url = "https://aclanthology.org/2025.findings-acl.239/",
    doi = "10.18653/v1/2025.findings-acl.239",
    pages = "4604--4621",
    ISBN = "979-8-89176-256-5"
}

@inproceedings{xu-etal-2025-alignment,
    title = "Alignment for Efficient Tool Calling of Large Language Models",
    author = "Xu, Hongshen  and
      Wang, Zihan  and
      Zhu, Zichen  and
      Pan, Lei  and
      Chen, Xingyu  and
      Fan, Shuai  and
      Chen, Lu  and
      Yu, Kai",
    editor = "Christodoulopoulos, Christos  and
      Chakraborty, Tanmoy  and
      Rose, Carolyn  and
      Peng, Violet",
    booktitle = "Proceedings of the 2025 Conference on Empirical Methods in Natural Language Processing",
    month = nov,
    year = "2025",
    address = "Suzhou, China",
    publisher = "Association for Computational Linguistics",
    url = "https://aclanthology.org/2025.emnlp-main.898/",
    doi = "10.18653/v1/2025.emnlp-main.898",
    pages = "17776--17792",
    ISBN = "979-8-89176-332-6"
}

@inproceedings{ICLR2026_788f0d33,
 author = {Chen, Yifei and Dong, Guanting and Dou, Zhicheng},
 booktitle = {International Conference on Learning Representations},
 editor = {C. Vondrick and B. Hariharan and C. Raffel and L. Pinto and D. Yang and A. Faust},
 pages = {74102--74122},
 title = {Toward Effective Tool-Integrated Reasoning via Self-Evolved Preference Learning},
 url = {https://proceedings.iclr.cc/paper_files/paper/2026/file/788f0d336eeddf698c8d527b1794fca4-Paper-Conference.pdf},
 volume = {2026},
 year = {2026}
}

@inproceedings{ICLR2026_85255cf8,
 author = {Wu, Peilin and Zhang, Mian and Wan, Kun and Zhao, Wentian and He, Kaiyu and Du, Xinya and Chen, Zhiyu},
 booktitle = {International Conference on Learning Representations},
 editor = {C. Vondrick and B. Hariharan and C. Raffel and L. Pinto and D. Yang and A. Faust},
 pages = {82317--82347},
 title = {HiPRAG: Hierarchical Process Rewards for Efficient Agentic Retrieval Augmented Generation},
 url = {https://proceedings.iclr.cc/paper_files/paper/2026/file/85255cf8f2d2318cc7cffd0b1bc66625-Paper-Conference.pdf},
 volume = {2026},
 year = {2026}
}

@inproceedings{qin-etal-2026-dora,
    title = "{DORA}: A Dual-Objective Reinforcement Learning Framework for Effective and Efficient Multimodal Agentic Search",
    author = "Qin, Guangming  and
      Deng, Yuhao  and
      Zhao, Yukun  and
      Li, Zhenyang  and
      Wang, Junfeng  and
      Yin, Dawei  and
      Yuan, Ye  and
      Wang, Guoren  and
      Yan, Yizhou  and
      Chai, Chengliang  and
      Cao, Lei",
    editor = "Liakata, Maria  and
      Moreira, Viviane P.  and
      Zhang, Jiajun  and
      Jurgens, David",
    booktitle = "Proceedings of the 64th Annual Meeting of the {A}ssociation for {C}omputational {L}inguistics (Volume 1: Long Papers)",
    month = jul,
    year = "2026",
    address = "San Diego, California, United States",
    publisher = "Association for Computational Linguistics",
    url = "https://aclanthology.org/2026.acl-long.605/",
    doi = "10.18653/v1/2026.acl-long.605",
    pages = "13252--13272",
    ISBN = "979-8-89176-390-6"
}

@Misc{vonwerra2020trl,
  title   = {{TRL: Transformers Reinforcement Learning}},
  author  = {von Werra, Leandro and Belkada, Younes and Tunstall, Lewis and Beeching, Edward and Thrush, Tristan and Lambert, Nathan and Huang, Shengyi and Rasul, Kashif and Gallouédec, Quentin},
  license = {Apache-2.0},
  howpublished = {\url{https://github.com/huggingface/trl}},
  url     = {https://github.com/huggingface/trl},
  year    = {2020}
}

@Misc{peft,
  title =        {{PEFT}: State-of-the-art Parameter-Efficient Fine-Tuning methods},
  author =       {Sourab Mangrulkar and Sylvain Gugger and Lysandre Debut and Younes Belkada and Sayak Paul and Benjamin Bossan and Marian Tietz},
  howpublished = {\url{https://github.com/huggingface/peft}},
  year =         {2022}
}

@inproceedings{wolf-etal-2020-transformers,
    title = "Transformers: State-of-the-Art Natural Language Processing",
    author = "Wolf, Thomas  and
      Debut, Lysandre  and
      Sanh, Victor  and
      Chaumond, Julien  and
      Delangue, Clement  and
      Moi, Anthony  and
      Cistac, Pierric  and
      Rault, Tim  and
      Louf, Remi  and
      Funtowicz, Morgan  and
      Davison, Joe  and
      Shleifer, Sam  and
      von Platen, Patrick  and
      Ma, Clara  and
      Jernite, Yacine  and
      Plu, Julien  and
      Xu, Canwen  and
      Le Scao, Teven  and
      Gugger, Sylvain  and
      Drame, Mariama  and
      Lhoest, Quentin  and
      Rush, Alexander",
    editor = "Liu, Qun  and
      Schlangen, David",
    booktitle = "Proceedings of the 2020 Conference on Empirical Methods in Natural Language Processing: System Demonstrations",
    month = oct,
    year = "2020",
    address = "Online",
    publisher = "Association for Computational Linguistics",
    url = "https://aclanthology.org/2020.emnlp-demos.6/",
    doi = "10.18653/v1/2020.emnlp-demos.6",
    pages = "38--45"
}

@inproceedings{3433701.3433727,
author = {Rajbhandari, Samyam and Rasley, Jeff and Ruwase, Olatunji and He, Yuxiong},
title = {ZeRO: memory optimizations toward training trillion parameter models},
year = {2020},
isbn = {9781728199986},
publisher = {IEEE Press},
booktitle = {Proceedings of the International Conference for High Performance Computing, Networking, Storage and Analysis},
articleno = {20},
numpages = {16},
location = {Atlanta, Georgia},
series = {SC '20}
}

@inproceedings{kwon2023efficient,
  title={Efficient Memory Management for Large Language Model Serving with PagedAttention},
  author={Woosuk Kwon and Zhuohan Li and Siyuan Zhuang and Ying Sheng and Lianmin Zheng and Cody Hao Yu and Joseph E. Gonzalez and Hao Zhang and Ion Stoica},
  booktitle={Proceedings of the ACM SIGOPS 29th Symposium on Operating Systems Principles},
  year={2023}
}

@inproceedings{
hu2022lora,
title={Lo{RA}: Low-Rank Adaptation of Large Language Models},
author={Edward J Hu and Yelong Shen and Phillip Wallis and Zeyuan Allen-Zhu and Yuanzhi Li and Shean Wang and Lu Wang and Weizhu Chen},
booktitle={International Conference on Learning Representations},
year={2022},
url={https://openreview.net/forum?id=nZeVKeeFYf9}
}

@misc{qwen3technicalreport,
      title={Qwen3 Technical Report}, 
      author={{Qwen Team}},
      year={2025},
      eprint={2505.09388},
      archivePrefix={arXiv},
      primaryClass={cs.CL},
      url={https://arxiv.org/abs/2505.09388}, 
}

@misc{liu2026ministral3,
      title={Ministral 3}, 
      author={Alexander H. Liu and Kartik Khandelwal and Sandeep Subramanian and Victor Jouault and Abhinav Rastogi and Adrien Sadé and Alan Jeffares and Albert Jiang and Alexandre Cahill and Alexandre Gavaudan and Alexandre Sablayrolles and Amélie Héliou and Amos You and Andy Ehrenberg and Andy Lo and Anton Eliseev and Antonia Calvi and Avinash Sooriyarachchi and Baptiste Bout and Baptiste Rozière and Baudouin De Monicault and Clémence Lanfranchi and Corentin Barreau and Cyprien Courtot and Daniele Grattarola and Darius Dabert and Diego de las Casas and Elliot Chane-Sane and Faruk Ahmed and Gabrielle Berrada and Gaëtan Ecrepont and Gauthier Guinet and Georgii Novikov and Guillaume Kunsch and Guillaume Lample and Guillaume Martin and Gunshi Gupta and Jan Ludziejewski and Jason Rute and Joachim Studnia and Jonas Amar and Joséphine Delas and Josselin Somerville Roberts and Karmesh Yadav and Khyathi Chandu and Kush Jain and Laurence Aitchison and Laurent Fainsin and Léonard Blier and Lingxiao Zhao and Louis Martin and Lucile Saulnier and Luyu Gao and Maarten Buyl and Margaret Jennings and Marie Pellat and Mark Prins and Mathieu Poirée and Mathilde Guillaumin and Matthieu Dinot and Matthieu Futeral and Maxime Darrin and Maximilian Augustin and Mia Chiquier and Michel Schimpf and Nathan Grinsztajn and Neha Gupta and Nikhil Raghuraman and Olivier Bousquet and Olivier Duchenne and Patricia Wang and Patrick von Platen and Paul Jacob and Paul Wambergue and Paula Kurylowicz and Pavankumar Reddy Muddireddy and Philomène Chagniot and Pierre Stock and Pravesh Agrawal and Quentin Torroba and Romain Sauvestre and Roman Soletskyi and Rupert Menneer and Sagar Vaze and Samuel Barry and Sanchit Gandhi and Siddhant Waghjale and Siddharth Gandhi and Soham Ghosh and Srijan Mishra and Sumukh Aithal and Szymon Antoniak and Teven Le Scao and Théo Cachet and Theo Simon Sorg and Thibaut Lavril and Thiziri Nait Saada and Thomas Chabal and Thomas Foubert and Thomas Robert and Thomas Wang and Tim Lawson and Tom Bewley and Tom Bewley and Tom Edwards and Umar Jamil and Umberto Tomasini and Valeriia Nemychnikova and Van Phung and Vincent Maladière and Virgile Richard and Wassim Bouaziz and Wen-Ding Li and William Marshall and Xinghui Li and Xinyu Yang and Yassine El Ouahidi and Yihan Wang and Yunhao Tang and Zaccharie Ramzi},
      year={2026},
      eprint={2601.08584},
      archivePrefix={arXiv},
      primaryClass={cs.CL},
      url={https://arxiv.org/abs/2601.08584}, 
}

@misc{qwen38,
    title = {{Qwen3.8-Max}: A New Bar for Coding and Cowork},
    url = {https://qwen.ai/blog?id=qwen3.8},
    howpublished = {\url{https://qwen.ai/blog?id=qwen3.8}},
    author = {{Qwen Team}},
    month = {August},
    year = {2026}
}
\appendix


\renewtcolorbox{promptbox}[2][]{enhanced, colback=black!2, colframe=black!78, coltitle=white,
  fonttitle=\small\bfseries, title={#2}, boxrule=0.6pt, arc=2pt,
  left=5pt, right=5pt, top=4pt, bottom=4pt, toptitle=2pt, bottomtitle=2pt, middle=4pt,
  segmentation style={black!40, dashed}, fontupper=\small\ttfamily, fontlower=\small\ttfamily,
  halign=left, halign lower=left, #1}
\renewcommand{\promptpart}[1]{{\normalfont\small\bfseries\color{black!60}#1}\par\smallskip}

\section{Experimental Details}
\label{app:setup}

\begin{table}[!b]
\centering
\small
\begin{tabular}{@{}lr@{}}
\toprule
\textbf{Hyperparameter} & \textbf{Value} \\
\midrule
\multicolumn{2}{@{}l}{\textit{Data synthesis}} \\
Intermediate answers per step $K$ & 3 \\
Temperature of intermediate answers & 0.6 \\
Maximum tokens of a rationale & 800 \\
\midrule
\multicolumn{2}{@{}l}{\textit{Fine-tuning}} \\
Halting weight $\lambda$ & 5 \\
LoRA rank / $\alpha$ / dropout & 32 / 64 / 0.1 \\
LoRA target modules & all linear layers \\
Optimizer & AdamW \\
Learning rate & $5 \times 10^{-4}$ \\
Learning rate schedule & cosine \\
Warmup ratio & 0.05 \\
Weight decay & 0.05 \\
Epochs & 4 \\
Effective batch size & 16 trajectories \\
Maximum length & 32,768 tokens \\
Precision & bf16 \\
\midrule
\multicolumn{2}{@{}l}{\textit{Evaluation}} \\
Temperature & 0 \\
Context window & 32,768 tokens \\
Step limit & 30 \\
\bottomrule
\end{tabular}
\caption{Hyperparameters shared by all runs. LoRA is applied to the linear layers of the language model.}
\label{tab:hyper}
\end{table}

\begin{figure*}[!t]
\begin{tcbraster}[raster columns=2, raster column skip=10pt, raster equal height=rows, raster before skip=0pt, raster after skip=8pt]
\begin{tcolorbox}[blanker]
\begin{promptbox}[height fill]{(a) Forced answer ($p_{\text{force}}$)}
\promptpart{System prompt}
You are an expert code analyst. Provide a final answer based on collected information.
\tcblower
\promptpart{User message}
-{}-{}- STOP AND ANSWER NOW -{}-{}-\\
This is the final allowed turn. You MUST NOT call any tools.\\
Provide your best possible final answer now inside a <finish>...</finish> block.\\[4pt]
Question: \ph{question}\\
Repository path: \ph{repository path}\\[4pt]
Chat history and tool results:\\
\ph{context $s_i$}
\end{promptbox}
\end{tcolorbox}
\begin{tcolorbox}[blanker]
\begin{promptbox}[height fill]{(b) Rationale generator}
\promptpart{System prompt}
You are helping build training data for a codebase question-answering agent.\\[4pt]
You will see a conversation in which the agent (the "assistant") explores a code repository with tools
to answer a user's question. The conversation stops at the point where the agent has gathered enough
evidence. You will then be given the final answer the agent is about to write.\\[4pt]
Your job: write the agent's last message before that final answer, in the agent's own voice.
The message should say what has been established (naming the specific files or functions seen in the
conversation) and that this is enough to answer the question. 2-4 sentences. Do not announce any further
investigation. Do not write the answer itself. Output only the message.\\[4pt]
Here are examples of how this agent writes that message:\\[4pt]
<example> \ph{example} </example> ($\times 3$)
\tcblower
\promptpart{Final user message}
The agent is now about to write this final answer:\\
<answer> \ph{answer} </answer>\\[4pt]
Write the agent's message right before this answer.
\end{promptbox}
\end{tcolorbox}
\end{tcbraster}
\caption{Prompts of JET. (a) Forced-answer instruction $p_{\text{force}}$, shown with the first line for golden step labeling and the autorater. (b) Prompt of the rationale generator for the rationale $r$.}
\label{fig:prompts-jet}
\end{figure*}

\subsection{Models}
The models in our experiments are as follows.
\begin{itemize}[leftmargin=*, itemsep=1pt, topsep=3pt, before=\raggedright]
\item \textbf{Agents.} \texttt{Qwen/Qwen3-8B} in the non-thinking mode and \texttt{mistralai/\allowbreak Ministral-3-8B-Instruct-2512-BF16}.
\item \textbf{Rationale generator.} \texttt{Qwen/Qwen3.8-27B} in the non-thinking mode at temperature 0.
\item \textbf{Labeling judge and autorater.} \texttt{gpt-6-luna} with medium reasoning effort.
\item \textbf{Evaluation judges.} \texttt{gpt-6-sol} with medium reasoning effort on SWE-QA-Pro and \texttt{claude-sonnet-5-5} on SWE-QA.
\end{itemize}
We accessed the three API models between September and October 2026.

\subsection{Training}
Table~\ref{tab:hyper} lists the hyperparameters, which all trained agents share.
We fine-tune the agents with TRL 1.12~\citep{vonwerra2020trl}, PEFT 0.20~\citep{peft}, and Transformers 5.8~\citep{wolf-etal-2020-transformers}, using DeepSpeed ZeRO-2~\citep{3433701.3433727}.
We serve the agents and the rationale generator with vLLM 0.20~\citep{kwon2023efficient} to generate the rollouts, the intermediate answers, and the rationales.
One of the 400 Qwen3 rollouts failed with a context-length error and is excluded from the training data.

\paragraph{Checkpoint selection.}
If the condition on tool calls in \S\ref{sec:evaluation} excludes every checkpoint, we select the checkpoint with the fewest tool calls.

\subsection{Evaluation}
\label{app:eval}

\paragraph{Cost.}
Cost in Table~\ref{tab:main-results} applies the OpenRouter price of the base model to every agent built on the base model, without cached-input pricing.
Qwen3-8B has no listed price, so both agent models use the price of Ministral-3-8B, 0.15 USD per million input and output tokens.
For the autorater, the price is 0.10 USD per million input tokens and 0.50 USD per million output tokens.

\paragraph{Failed questions.}
The SWE-QA-Pro scaffold restarts a question after an error, for up to seven attempts, and the judge scores the error message of a question that still fails.
We count Calls and Cost on the last attempt only.
For a failed question, we set Calls to the maximum of 29 tool calls and exclude the question from Cost.
Restarts and failures occur mostly with the SFT agent, so both rules favor the SFT baseline rather than JET.

\begin{figure*}[!t]
\begin{tcbraster}[raster columns=2, raster column skip=10pt, raster equal height=rows, raster before skip=0pt, raster after skip=8pt]
\begin{tcolorbox}[blanker]
\begin{promptbox}[height fill]{(a) Budget Tracker}
\promptpart{Guideline appended to the system prompt}
\#\# Budget\\[4pt]
You have one tool-call budget of \ph{budget} tool calls. Each tool call consumes 1 unit.\\[4pt]
After each tool response, a <budget> tag shows remaining units. You must ADAPT your strategy dynamically to the current budget state.\\[4pt]
\#\#\# HIGH Budget (>=70\% remaining)\\
- Search: 3-5 diverse semantic\_search or grep queries in one turn.\\
- View: up to 2-3 high-value files or line ranges.\\
- Goal: Broad exploration, build context fast.\\[4pt]
\#\#\# MEDIUM Budget (30\%-70\%)\\
- Search: 2-3 precise, refined queries per turn.\\
- View: 1-2 files or line ranges that close key knowledge gaps.\\
- Goal: Converge; eliminate uncertainty efficiently.\\[4pt]
\#\#\# LOW Budget (10\%-30\%)\\
- Search: 1 tightly focused query.\\
- View: at most 1 most promising file or line range.\\
- Goal: Verify a single critical fact or finalize answer.\\[4pt]
\#\#\# CRITICAL (<10\% remaining)\\
- Only perform 1 minimal-cost search or view if absolutely essential.\\
- If uncertainty remains and no tool use is possible, give your best answer in the <finish> block.\\[4pt]
Always justify tool choices based on the remaining budget.
\tcblower
\promptpart{Block appended after the question and every step}
<budget>\\
Tool Call Budget Used: \ph{used}, Tool Call Budget Remaining: \ph{remaining}\\
Make the best use of the available resources.\\
</budget>
\end{promptbox}
\end{tcolorbox}
\begin{tcolorbox}[blanker]
\begin{promptbox}[height fill]{(b) Autorater}
You are an expert LLM evaluator that excels at evaluating a QUESTION and REFERENCES.
Consider the following criteria:\\
Sufficient Context: 1 IF the CONTEXT is sufficient to infer the answer to the question and 0 IF the CONTEXT cannot be used to infer the answer to the question\\[4pt]
First, output a list of step-by-step questions that would be used to arrive at a label for the criteria. Make sure to include questions about assumptions implicit in the QUESTION. Include questions about any mathematical calculations or arithmetic that would be required. Next, answer each of the questions. Make sure to work step by step through any required mathematical calculations or arithmetic. Finally, use these answers to evaluate the criteria.\\[4pt]
Output the \#\#\# EXPLANATION (Text). Then, use the EXPLANATION to output the \#\#\# EVALUATION (JSON)\\[4pt]
EXAMPLE:\\
\ph{example}\\[4pt]
Remember the instructions: [the two paragraphs above, repeated]\\[4pt]
\#\#\# QUESTION\\
\ph{question}\\
\#\#\# REFERENCES\\
\ph{context}
\end{promptbox}
\end{tcolorbox}
\end{tcbraster}
\caption{Prompts of the baselines. (a) Prompt of Budget Tracker, adapted from \citet{liu2026budgetaware}. (b) Prompt of the autorater, adapted from \citet{ICLR2025_33dffa2e}.}
\label{fig:prompts-baselines}
\end{figure*}

\section{Prompts}
\label{app:prompts}

Figure~\ref{fig:prompts-jet} shows the prompts of JET, and Figure~\ref{fig:prompts-baselines} shows the prompts of the baselines.
Red text in braces marks a placeholder.

\paragraph{Forced answer.}
Figure~\ref{fig:prompts-jet}a shows the forced-answer instruction $p_{\text{force}}$, which follows the forced-answer prompt of the SWE-QA-Pro scaffold.
The scaffold uses \texttt{MAX TURNS REACHED} as the first line at the step limit, and we replace this line with \texttt{STOP AND ANSWER NOW} for golden step labeling and the autorater.

\paragraph{Rationale.}
Figure~\ref{fig:prompts-jet}b shows the prompt of the rationale generator.
After the system prompt, we provide the context $s_{i^\star}$.
We then provide the golden step answer $\hat{a}$ in a final user message and ask the rationale generator to write the rationale.
The system prompt contains three examples, which are the messages that Claude Sonnet 4.5 wrote before the final answer in three of the expert trajectories.
We use the same examples in the Self and Expert settings.

\begin{table*}[t]
\centering
\small
\setlength{\aboverulesep}{0pt}
\setlength{\belowrulesep}{0pt}
\setlength{\tabcolsep}{3.2pt}
\renewcommand{\arraystretch}{1.15}
\begin{tabular}{llcccccc!{\color{black!35}\vrule}cccccc}
\toprule
\textbf{Training data} & \textbf{Method} & \multicolumn{6}{c!{\color{black!35}\vrule}}{\textbf{SWE-QA-Pro}} & \multicolumn{6}{c}{\textbf{SWE-QA}} \\
 & & Corr. & Comp. & Rel. & Clar. & Reas. & Score & Corr. & Comp. & Rel. & Clar. & Reas. & Score \\
\midrule
\rowcolor{tabhead}\multicolumn{14}{l}{\textbf{Qwen3-8B}} \\
\multirow{5}{*}{Self} & Base & 4.77 & 3.78 & 8.14 & 8.77 & 5.42 & 30.88 & 6.25 & 4.40 & 8.98 & 11.13 & 5.74 & 36.50 \\
 & Budget Tracker & 4.47 & 3.38 & 8.59 & 8.73 & 5.11 & 30.29 & 6.02 & 3.92 & 8.87 & 10.79 & 5.29 & 34.88 \\
 & Autorater & 4.81 & 3.63 & 8.23 & 8.78 & 5.25 & 30.70 & 6.19 & 4.09 & 8.96 & 10.90 & 5.46 & 35.61 \\
 & Trunc. SFT & 4.92 & 3.63 & 8.62 & 8.77 & 5.28 & 31.22 & 6.48 & 4.23 & 9.37 & 11.20 & 5.63 & 36.91 \\
 & \textbf{JET} & 4.64 & 3.65 & 8.37 & 8.76 & 5.43 & 30.84 & 6.10 & 4.24 & 9.02 & 11.23 & 5.75 & 36.34 \\
\arrayrulecolor{black!35}\specialrule{0.4pt}{0pt}{0pt}\arrayrulecolor{black}
\multirow{5}{*}{Expert} & SFT & 6.40 & 6.31 & 8.45 & 8.84 & 7.08 & 37.07 & 9.39 & 8.90 & 12.02 & 13.24 & 9.24 & 52.79 \\
 & Budget Tracker & 6.37 & 6.30 & 8.53 & 8.84 & 7.05 & 37.08 & 9.45 & 9.02 & 12.03 & 13.27 & 9.33 & 53.09 \\
 & Autorater & 5.52 & 5.17 & 8.63 & 8.84 & 6.33 & 34.49 & 8.44 & 7.58 & 11.48 & 12.69 & 8.24 & 48.42 \\
 & Trunc. SFT & 6.24 & 6.05 & 8.45 & 8.85 & 6.93 & 36.52 & 9.19 & 8.69 & 11.91 & 13.04 & 9.01 & 51.83 \\
 & \textbf{JET} & 6.13 & 5.35 & 8.84 & 8.96 & 6.65 & 35.93 & 8.71 & 6.69 & 11.58 & 12.74 & 7.75 & 47.47 \\
\arrayrulecolor{black!35}\specialrule{0.4pt}{0pt}{0pt}\arrayrulecolor{black}
\multirow{2}{*}{Expert $\rightarrow$ Self} & \textbf{JET} $\rightarrow$ Trunc. SFT & 5.93 & 5.28 & 8.68 & 8.90 & 6.50 & 35.29 & 8.38 & 6.72 & 11.20 & 12.61 & 7.68 & 46.59 \\
 & \textbf{JET} $\rightarrow$ \textbf{JET} & 5.90 & 5.23 & 8.58 & 8.88 & 6.50 & 35.08 & 8.39 & 6.45 & 11.26 & 12.55 & 7.60 & 46.25 \\
\midrule
\rowcolor{tabhead}\multicolumn{14}{l}{\textbf{Ministral-3-8B}} \\
\multirow{5}{*}{Self} & Base & 6.20 & 5.12 & 9.02 & 9.00 & 6.50 & 35.84 & 8.76 & 6.21 & 11.46 & 12.54 & 7.31 & 46.27 \\
 & Budget Tracker & 6.20 & 5.08 & 9.05 & 8.94 & 6.47 & 35.75 & 8.87 & 6.14 & 11.56 & 12.58 & 7.24 & 46.38 \\
 & Autorater & 5.54 & 4.90 & 8.64 & 8.77 & 6.11 & 33.96 & 8.07 & 6.78 & 10.95 & 12.43 & 7.70 & 45.93 \\
 & Trunc. SFT & 6.07 & 5.15 & 8.85 & 8.93 & 6.52 & 35.53 & 8.58 & 6.42 & 11.34 & 12.52 & 7.46 & 46.32 \\
 & \textbf{JET} & 5.82 & 5.23 & 8.53 & 8.81 & 6.42 & 34.81 & 8.21 & 6.79 & 10.86 & 12.43 & 7.67 & 45.97 \\
\arrayrulecolor{black!35}\specialrule{0.4pt}{0pt}{0pt}\arrayrulecolor{black}
\multirow{5}{*}{Expert} & SFT & 6.06 & 6.05 & 7.90 & 8.46 & 6.64 & 35.10 & 9.31 & 9.06 & 11.52 & 12.71 & 9.18 & 51.78 \\
 & Budget Tracker & 6.22 & 6.15 & 8.13 & 8.60 & 6.84 & 35.94 & 9.38 & 9.10 & 11.73 & 12.93 & 9.24 & 52.37 \\
 & Autorater & 5.65 & 5.55 & 8.34 & 8.53 & 6.41 & 34.48 & 8.98 & 8.34 & 11.73 & 12.83 & 8.77 & 50.64 \\
 & Trunc. SFT & 6.42 & 6.35 & 8.30 & 8.87 & 7.08 & 37.02 & 9.82 & 9.28 & 12.06 & 13.45 & 9.61 & 54.22 \\
 & \textbf{JET} & 6.38 & 6.06 & 8.54 & 8.88 & 7.04 & 36.91 & 9.70 & 8.88 & 12.26 & 13.53 & 9.47 & 53.83 \\
\arrayrulecolor{black!35}\specialrule{0.4pt}{0pt}{0pt}\arrayrulecolor{black}
\multirow{2}{*}{Expert $\rightarrow$ Self} & \textbf{JET} $\rightarrow$ Trunc. SFT & 6.09 & 5.76 & 8.30 & 8.76 & 6.76 & 35.67 & 8.93 & 8.22 & 11.61 & 13.00 & 8.73 & 50.49 \\
 & \textbf{JET} $\rightarrow$ \textbf{JET} & 5.71 & 5.41 & 8.37 & 8.70 & 6.44 & 34.63 & 8.54 & 7.66 & 11.39 & 12.68 & 8.29 & 48.55 \\
\bottomrule
\end{tabular}
\caption{Mean score of each judge criterion. Each criterion ranges from 1 to 10 on SWE-QA-Pro and from 1 to 20 on SWE-QA, and Score is the sum of the five criteria, as reported in Table~\ref{tab:main-results}.}
\label{tab:criteria}
\end{table*}

\paragraph{Budget Tracker.}
We append the guideline in Figure~\ref{fig:prompts-baselines}a to the system prompt of the agent.
The guideline follows the prompt of \citet{liu2026budgetaware}, in which we merge the two tool budgets into one and replace the tools and the answer tag with those of our agent.
After the question and after every step, the scaffold appends the budget block.

\paragraph{Autorater.}
Figure~\ref{fig:prompts-baselines}b shows the prompt of the autorater, which follows the sufficient-context prompt of \citet{ICLR2025_33dffa2e}.
We provide the current context of the agent as REFERENCES.
We drop the timestamp line and replace the example with a rollout of the base agent on a training question of SWE-QA-Pro.
Once the autorater outputs \texttt{"Sufficient Context": 1}, the scaffold forces the agent to answer.

\section{Additional Results}
\label{app:results}

\paragraph{Judge criteria.}
Table~\ref{tab:criteria} breaks down the scores in Table~\ref{tab:main-results} by judge criterion.
The five criteria are correctness, completeness, relevance, clarity, and reasoning.

\end{document}